\documentclass[sigconf]{acmart}
\AtBeginDocument{%
  \providecommand\BibTeX{{%
    \normalfont B\kern-0.5em{\scshape i\kern-0.25em b}\kern-0.8em\TeX}}}

\copyrightyear{2023}
\acmYear{2023}
\acmConference[KDD FL4Data-Mining '23]{KDD FL4Data-Mining '23: International Workshop on Federated Learning for Distributed Data Mining}{August 7, 2023}{Long Beach, CA, USA}
\acmBooktitle{KDD FL4Data-Mining '23, August 7, 2023, Long Beach, CA, USA.}

\begin{document}

\title{Asynchronous Decentralized Federated Lifelong Learning for Landmark Localization in Medical Imaging}

\author{Guangyao Zheng}
\affiliation{%
  \institution{Rice University}
  \streetaddress{6100 Main St}
  \city{Houston}
  \state{Texas}
  \country{USA}
  \postcode{77005}
}
\email{tz30@rice.edu}

\author{Vladimir Braverman}
\affiliation{%
  \institution{Rice University}
  \streetaddress{6100 Main St}
  \city{Houston}
  \state{Texas}
  \country{USA}}

\author{Michael A. Jacobs}
\affiliation{%
  \institution{Department Of Diagnostic And Interventional Imaging, McGovern Medical School, UTHealth}
  \streetaddress{6431 Fannin St}
  \city{Houston}
  \state{Texas}
  \country{USA}
  \postcode{77030}
}

\author{Vishwa S. Parekh}
\affiliation{%
 \institution{The University of Maryland Medical Intelligent Imaging (UM2ii) Center}
 \streetaddress{520 W Lombard St}
 \city{Baltimore}
 \state{Maryland}
 \country{USA}}
 \postcode{21201}

\renewcommand{\shortauthors}{Zheng, et al.}

\begin{abstract}
  Federated learning is a recent development in the machine learning area that allows a system of devices to train on one or more tasks without sharing their data to a single location or device. However, this framework still requires a centralized server to consolidate individual models into one synchronously or have inefficient or frail peer-to-peer communication, which are potential bottlenecks for the use of federated learning. In this paper, we propose a novel method of asynchronous decentralized federated lifelong learning (ADFLL) method that inherits the merits of federated learning and can train on multiple tasks simultaneously without the need for a central node or synchronous training, or less-than-desirable peer-to-peer communication. Thus, overcoming the potential drawbacks of conventional federated learning.  We demonstrate excellent performance on the brain tumor segmentation (BRATS) dataset for localizing the left ventricle on multiple image sequences and image orientation. Our framework allows agents to achieve the best performance with a mean distance error of 7.81, better than the conventional central aggregation agent's mean distance error of 11.78, and significantly (p=0.01) better than a conventional lifelong reinforcement learning (LL) agent with a distance error of 15.17 after eight rounds of training. In addition, all ADFLL agents have better performance than a conventional reinforcement learning (RL) agent with no LL implementation. In conclusion, we developed an ADFLL framework with excellent performance and speed-up compared to conventional LL agents.

\end{abstract}


\received{20 February 2007}
\received[revised]{12 March 2009}
\received[accepted]{5 June 2009}

\maketitle

\section{Introduction}
Medical imaging, MRI (Magnetic Resonance Imaging), PET (Positron Emission Tomography), CT (Computerized Tomography), X-ray, and Ultrasound, play a critical role in the diagnosis, prognosis, and preventative care of patients. The use of machine learning methods in medical imaging, such as classification, segmentation, noise reduction, and landmark localization, has been used in different completing complicated environments and settings \cite{CHENG2022102313,PAN2022103824,doi:10.1080/02564602.2021.1937349,KHAIRANDISH2022290,9139480}. However, these methods are usually done on single tasks, without the ability to generalize to other tasks. They often require a full dataset on a device for training, which may cause privacy concerns about patient data and computational constraints for the device specifications \cite{10.1007/978-3-030-91387-8_1}.

To address these challenges, Federated Learning (FL) has emerged as a promising approach that enables multiple agents to collaboratively train a model without sharing their data \cite{97c9f5c0c4714251a5c377616bf32211}. A federated learning system aims to protect data privacy and reduce computational costs at the local agent level by distributing the computation to multiple agents to train the model on their local data and sharing only the model updates with a central server. Federated learning implementations have shown promising results in various medical applications \cite{10.1007/978-3-030-60548-3_18,Jiang_Wang_Dou_2022,9268161}. However, federated learning frameworks often rely on synchronized learning schedules, meaning all participating agents start training at the same time. They also require agents to have the same architecture in order for the central server to aggregate the model weights. Data and agent heterogeneity influence the training speed which greatly reduces the efficiency and challenges the robustness of these approaches \cite{58f2965c5c4847d8b5e02e9e4408799d}. Additionally, Federated learning approaches cannot perform Lifelong Learning (LL), which is an important aspect of machine learning applied to medical imaging.  Works have shown the ability to improve accuracy, have excellent performance on multiple tasks, and generalize \cite{10.1007/978-3-030-00928-1_54}. With medical imaging tasks involving constant and rapid altering in imaging environments, such as new imaging sequences or abnormal patient conditions, a federated learning framework that is trained in an older environment may not perform well when evaluated in an unseen environment. One way to address this issue is to train the model again on the new dataset. However, new environment data can potentially be scarce, and this approach may lead to catastrophic forgetting, where the model loses its ability to operate effectively in the older environment. Federated lifelong learning implementations such as Huang et. al. exist \cite{2204.13591} in the medical field, but it is limited by their ring-type structure, which means the communication delay can potentially be very high due to irresponsive or failed agents that bottleneck the entire system.

To address all the limitations mentioned above, we propose an asynchronous decentralized federated lifelong learning (ADFLL) approach to landmark localization in medical imaging. This framework leverages Federated learning's ability to protect data privacy and reduce computational constraints, while also permitting data and agent heterogeneity to be in the system. This framework does not require a central node, and can together lifelong learn multiple tasks without catastrophic forgetting. We provide a flexible, efficient, and robust framework that can be deployed in real-world applications. This paper presents experimental results demonstrating the efficacy of our framework on the 2017 brain tumor segmentation (BraTS) dataset consisting of 8 different image environments and imaging sequences, highlighting its potential to revolutionize landmark localization in the medical imaging field while also maintaining data privacy and reducing computational costs.

\begin{figure}[!htb]\centering{}
\includegraphics[width=0.5\textwidth]{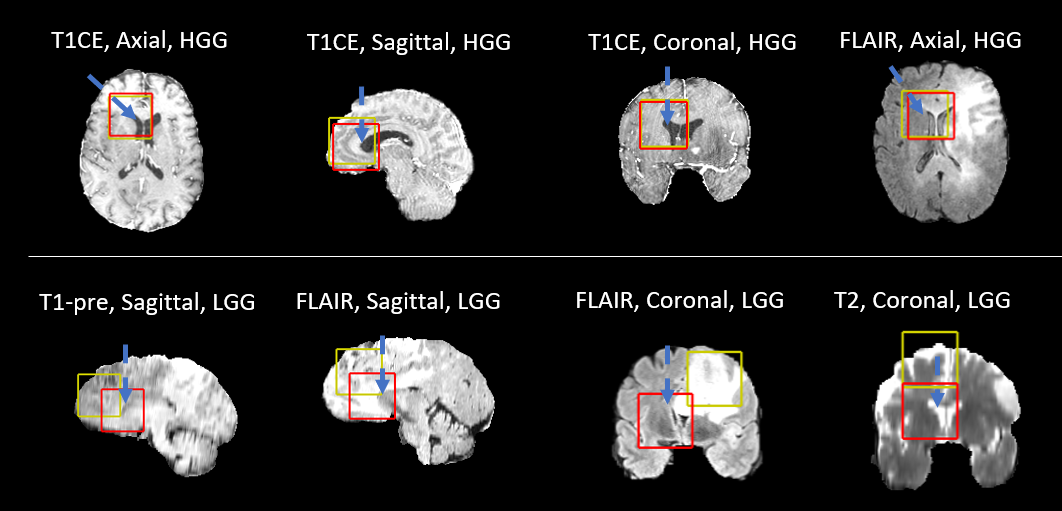}
\caption{Illustration of the 8 task-environment pairs. The red boxes indicate the true landmark location of the top left ventricle. The yellow box is a predicted location from ADFLL agents during their training progressions} \label{fig3}
\end{figure}

\section{Experiment and Result}

\subsection{Dataset and Experimental Setup}
\subsubsection{Clinical Data}
For evaluation of our ADFLL  framework, we utilized the 2017 brain tumor segmentation (BraTS) dataset \cite{menze2014multimodal}. This dataset consisted of 285 patients and included pre-contrast T1-weight, post-contrast T1-weighted, T2-weighted, and Fluid Attenuated Inversion Recovery (FLAIR) sequences in the axial orientation. From this dataset, we randomly sampled a subset of 100 patients to use as our experiment dataset. 60 patients have high-grade glioma (HGG) and 40 patients have low-grade glioma (LGG). We split the 100 patients into two parts 80:20, 80 were used for training and 20 for evaluation, with the training set consisting of 48 HGG and 32 LGG tumors, and the test set consisting of 12 HGG and 8 LGG tumors. We reconstructed the dataset to include all three imaging orientations (coronal, sagittal, and axial). As a result, we obtained a total of twenty-four imaging environments with combinations of two pathologies, 4 imaging sequences, and 3 image orientations. The top left ventricle was chosen as the task for this experiment, and 8 task-environment pairs were sampled as shown in Fig.~\ref{fig3}.

\subsubsection{Deployment Experimental Setup}

\begin{figure}
\centering{}
\includegraphics[width=0.4\textwidth]{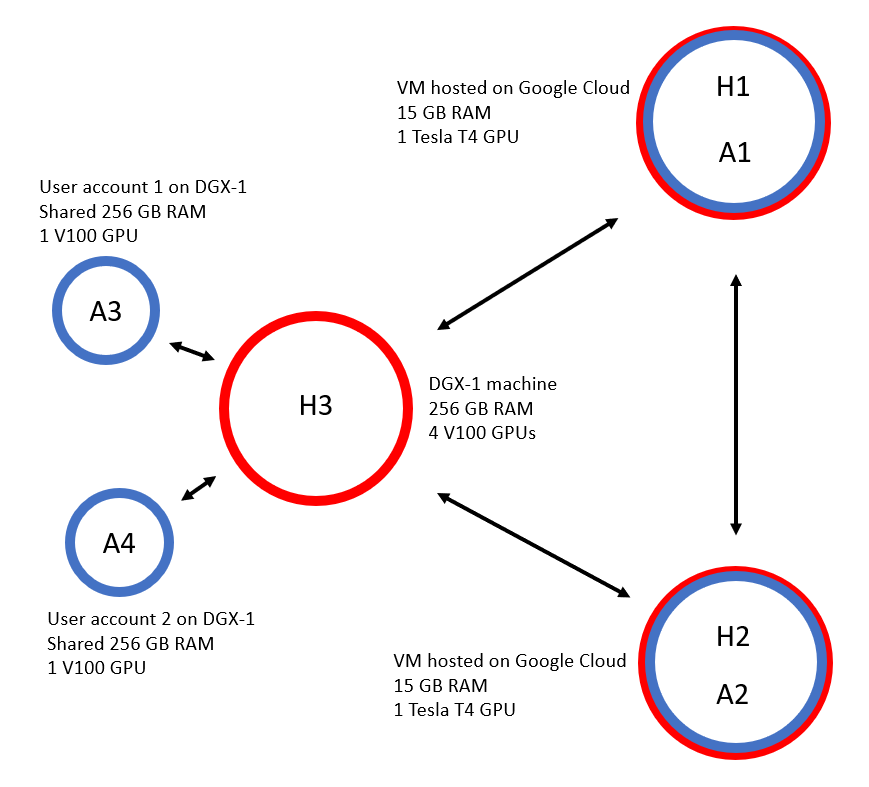}
\caption{Illustration of the 4-agent decentralized federated lifelong learning framework of our experiment.} \label{fig4}
\end{figure}

Every agent implements a multi-task lifelong reinforcement learning algorithm for localizing landmarks across the human anatomy. The federated lifelong learning component is implemented by integrating experience replay buffers from previous experiences shared by agents across the network for training. There are four agents in this experimental system: we implemented two on an NVIDIA DGX-1 each with an NVIDIA V100 and two on Google Cloud each with an NVIDIA T4. The topology of the system is shown in Fig.~\ref{fig4}. The two agents A1, and A2 on Google Cloud have their individual hubs H1 and H2. The two agents A3, and A4 on the DGX-1 are connected to the third hub H3 with a total of three hubs for 4 agents. Since the GPUs on DGX-1 are much more powerful than the GPUs on Google Cloud, A3 and A4 will run significantly faster than A1 and A2. We also implemented asynchronous learning, meaning when the agent finishes training on a task, as long as there are new ERBs that they have not learned from, they will start a new round and learn from those ERBs. Each agent will also get a different image training dataset each round. This process is continued until all four agents complete three rounds of training, guaranteeing all 8 sampled tasks for this experiment will be learned by the system.

\begin{table*}[htb]
\includegraphics[width=\textwidth]{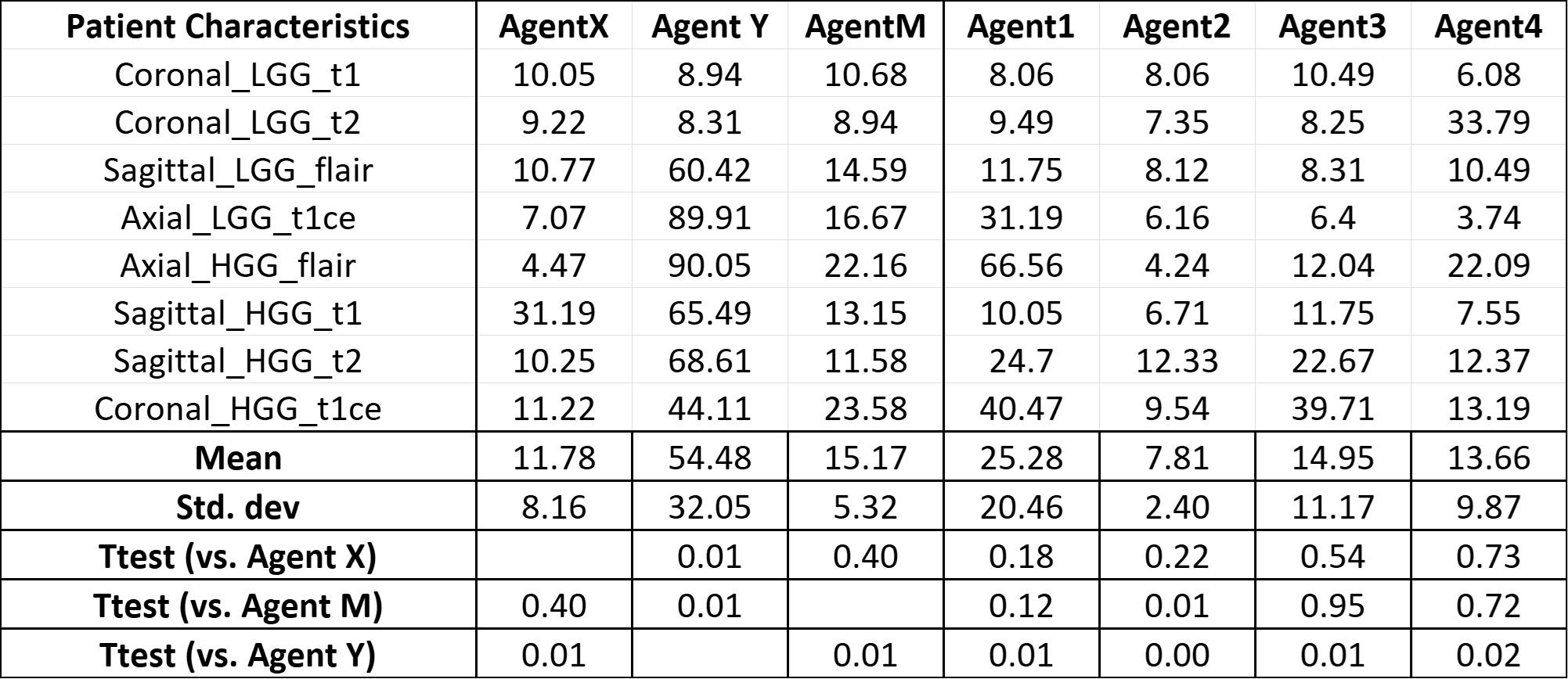}
\caption{Comparison of distance error between our agents (Agent 1-4) after round 3, all-knowing deep reinforcement learning agent (Agent X) after round 1, partially-knowing deep reinforcement learning agent (Agent Y) after round 1, and traditional lifelong deep reinforcement learning agent (Agent M) after round 8.}\label{fig5} 
\end{table*}

\textbf{All-knowing agent and partially-knowing agent:} To better compare our framework with non-lifelong learning ones, we ran two different deep reinforcement learning agents. Agent X is the all-knowing agent, with all 8 datasets available to it at the start of training for the deployment experiment and 24 datasets for the ablation study. It trained on the available data for one round. Agent Y is the partially-knowing agent, which only has access to one dataset and can therefore only train for one round.

\textbf{Traditional lifelong deep reinforcement learning agent:}
To better compare our framework with the traditional lifelong learning framework, we ran an Agent M that has access to the dataset sequentially and is therefore trained for eight rounds to account for learning eight different environments for the deployment experiment and two rounds for the ablation study.

\textbf{Experimental Metric:}
The performance metric was set as the terminal Euclidean distance between the agent's prediction and the target landmark. We performed paired t-tests to compare the performance of the decentralized federated lifelong learning framework with the traditional lifelong learning framework and all-knowing deep reinforcement learning agent and partial-knowing deep reinforcement learning agent. The p-value for statistical significance was set to $p \le 0.05$. 
\subsubsection{Ablation Study}

We conducted two simulation experiments to evaluate the scalability, flexibility, and robustness of our framework. We initialized the same type of agents in both experiments as in the functionality experiment. For both experiments, we evaluated the average performance of all the agents for the task of localizing the top left ventricle across all 24 imaging environments. Additionally, since it is prohibitively expensive to experiment on 24 different machines, these systems were simulated on the NVIDIA DGX-1, with a synchronous training protocol.

\textbf{Addition of agents experiment:}
We initialized a system with four agents, as previously described in the functionality experiment. We subsequently increased the number of agents in the system from 4 to 16 agents over the progression of 4 rounds (4,8,12,16). We further simulated a communication dropout of $75\%$ to account for network communication issues in the real world leading information loss while transmitting ERBs across agents. The goal of this experiment was to demonstrate how newer agents joining the system at different points in time can take advantage of the available within the system to learn the collective knowledge available in the system within just one round. 

\textbf{Deletion of agents experiment:}
In the deletion experiment, we gradually decreased The number of agents in the system from 24 to 1 agent over the progression of 5 rounds (24,12,6,3,1). The communication for this experiment was also simulated with a $75\%$ dropout. The goal of this experiment was to demonstrate how the proposed ADFLL system preserves the collective knowledge in a lifelong learning manner across all the tasks even as the agents contributing the knowledge leave the system.

\subsection{Results}
We conducted a deployment experiment based on 8 sub-task-environment pairs: Axial HGG t1ce, Sagittal HGG t1ce, Coronal HGG t1ce, Axial HGG flair, Sagittal LGG flair, Coronal LGG flair, Coronal LGG t2, Sagittal LGG t1. We sampled one image from each task to test the performance of our model and baseline models. Each round the four federated lifelong learning will receive a new task. They will begin the next round when there is also ERB to train from. Since the agents' training speeds are very different, A1 and A2 will finish their tasks slower, allowing them to learn from more ERBs at once. As shown in Table \ref{fig5}, after three rounds of training, A2 was able to achieve a mean distance error of 7.81 on all 8 tasks, compared to the all-knowing agent's 11.78 (p=0.22), but significantly lower compared to partially-knowing agent's 54.58 (p<0.001), and the traditional lifelong learning agent's 15.17 (p=0.01) after eight rounds of training. Since this is a real-world experiment, other agents trained faster than A2, meaning that they did not have all the ERBs available to them when they started their last round of training, resulting in their performances being worse than A2. The possible reasons for Agent 1 to have lower performance than Agent X and Agent M can be because it did not have all ERBs or the training was stuck at a local minimum. This can be easily solved by sharing the model parameters of the latest agent. Note that the all-knowing agent and the partially-knowing agent only train for 1 round for this experiment because they have no lifelong learning capability. 

\begin{figure*}[htb]
\includegraphics[width=\textwidth]{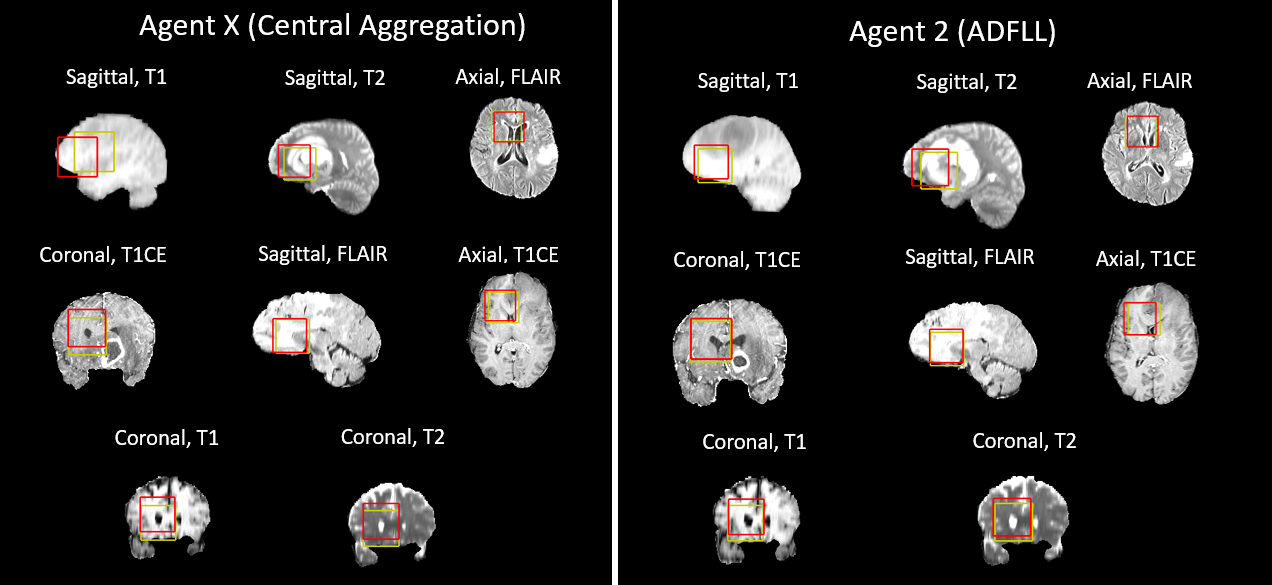}
\caption{Comparison of distance error across 8 different tasks between Agent X (Central Aggregation) and Agent 2 (ADFLL).} \label{fig6}
\end{figure*}

Compared to the All-knowing agent shown in \ref{fig6}, our framework is able to achieve an excellent performance boost. Compared to the traditional lifelong learning agent, our framework is able to achieve a significant performance boost and an outstanding speedup.

\begin{figure}[htb!]
\includegraphics[width=0.5\textwidth]{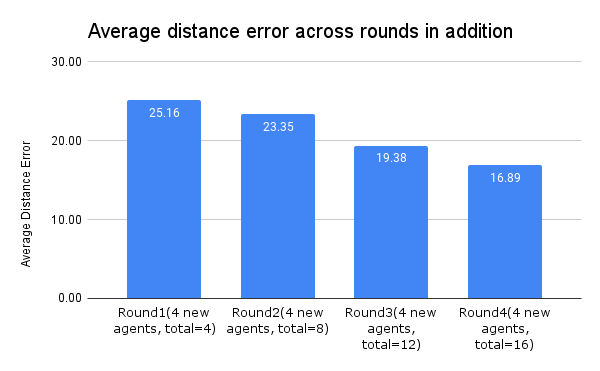}
\caption{Comparison of distance error of all agents in the system across 4 rounds of training} \label{fig10}
\end{figure}

\begin{figure}[htb!]
\includegraphics[width=0.5\textwidth]{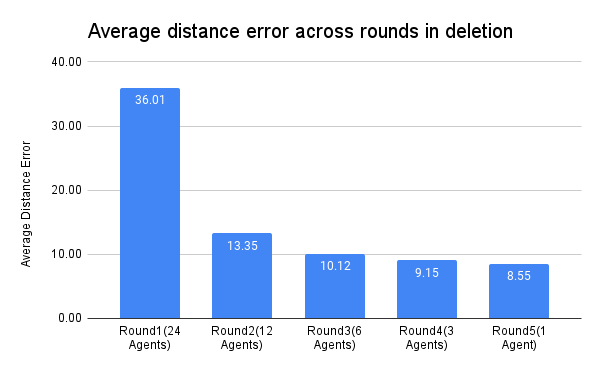}
\caption{Comparison of distance error of all agents in the system across 4 rounds of training} \label{fig11}
\end{figure}

In our two ablation studies, our framework showed scalability of up to 24 agents, robustness against network dropout, and flexibility in system topology. As shown in Figure \ref{fig10}, we see that the average Euclidean distance error across all agents decreases as more agents are added to the system, with an average Euclidean distance error of 16.89 at the end of 4 rounds. As shown in Figure \ref{fig11}, we also see that the average Euclidean distance error across all agents decreases, while half of the agents are deleted every round, resulting in an average Euclidean distance error of 8.55 after 5 rounds. This shows that the knowledge agents learned, captured in ERBs are not lost when agents are removed from the system. And when agents are being added, the new agents can catch up to existing agents in one round. Moreover, the $75\%$ dropout rate that is applied to every round of both experiments shows the robustness of our framework against network failures, a major bottleneck for federated learning frameworks.

\section{Conclusion}
Previous works have explored the application of federated learning to the medical field \cite{PPR:PPR463670,electronics11101548,Roy2019BrainTorrentAP}. They have shown decentralized federated learning system setups, each demonstrating good performance in their experiment tasks. But because of their system topology implementation, one node failure can potentially collapse the entire system. Moreover, the learning tasks examined were limited, binary classifications or MNIST dataset classification, resulting in limited potential applications. Additionally, their implementation offers a synchronous training procedure, which means in a real application scenario, users of their framework will have to coordinate the training process.

Asynchronous federated learning has also been explored in other areas \cite{Chen2019AsynchronousOF}. They offer the ability to deal with nodes with different computational power but lack the decentralization that allows the system to be more flexible.

Asynchronous decentralized federated learning has also been explored \cite{Liu2022AsynchronousDF,2204.13591}. However, they are still limited in their system implementation. The cost of removing a central node is a quadratic complexity communication scheme in that every node communicates with every node.

We have demonstrated a privacy-aware, asynchronous decentralized federated learning system with robust and efficient system topology. We have demonstrated excellent performance on landmark localization tasks on the BraTS image dataset. Our framework performs better than all-knowing deep reinforcement learning agents and traditional lifelong learning agents. Moreover, in our ablation study, our framework demonstrated excellent scalability, flexibility, and robustness. In the future, we will optimize our approach, expand the system further, and increase computational efficiency.
\begin{figure*}\centering{}[htb]
\includegraphics[width=0.8\textwidth]{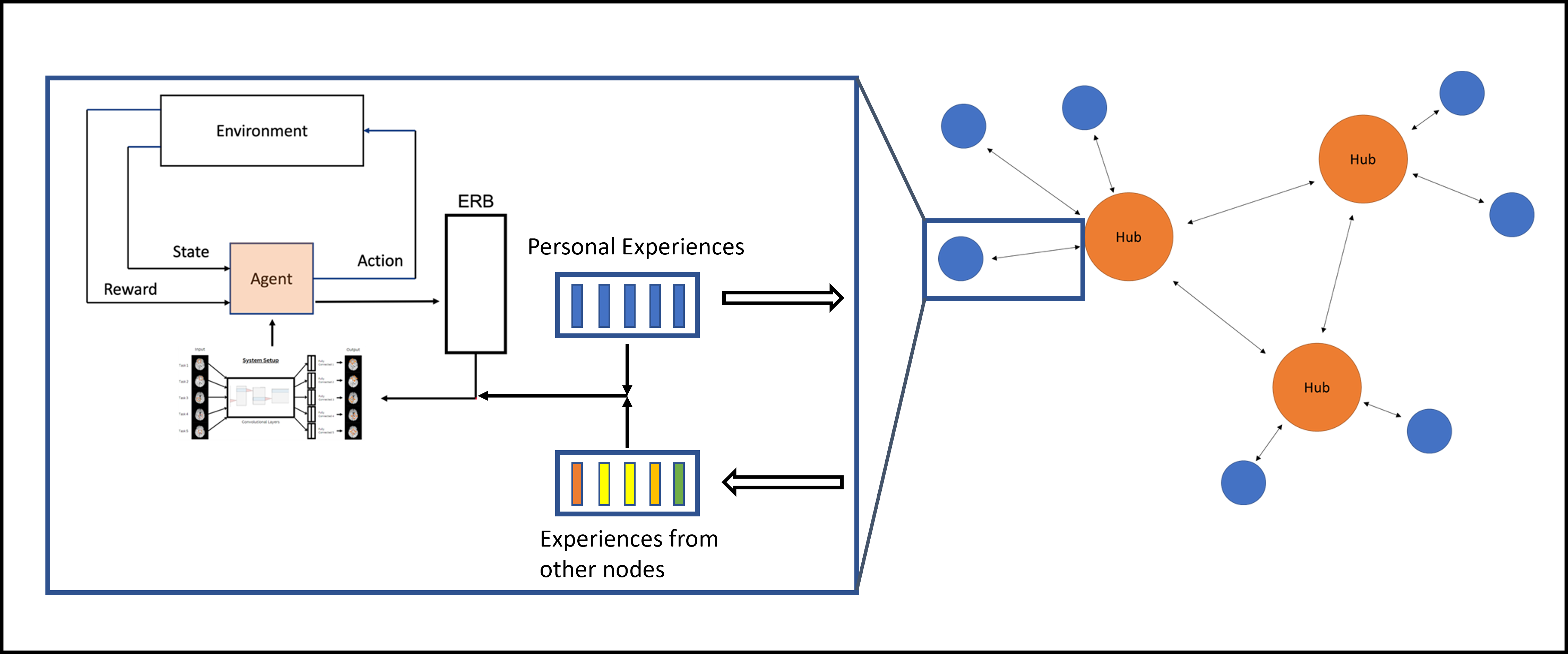}
\caption{Illustration of decentralized federated learning setup. Blue circles represent individual agents, and orange circles represent hubs.} \label{fig2}
\end{figure*}

\begin{acks}
This work was supported by the DARPA grant: DARPA-PA-20-02-11-HR00112190130  and 5P30CA006973 (Imaging Response Assessment Team-IRAT), U01CA140204
\end{acks}

\bibliographystyle{ACM-Reference-Format}
\bibliography{sample-base}


\begin{thebibliography}{24}


\ifx \showCODEN    \undefined \def \showCODEN     #1{\unskip}     \fi
\ifx \showDOI      \undefined \def \showDOI       #1{#1}\fi
\ifx \showISBNx    \undefined \def \showISBNx     #1{\unskip}     \fi
\ifx \showISBNxiii \undefined \def \showISBNxiii  #1{\unskip}     \fi
\ifx \showISSN     \undefined \def \showISSN      #1{\unskip}     \fi
\ifx \showLCCN     \undefined \def \showLCCN      #1{\unskip}     \fi
\ifx \shownote     \undefined \def \shownote      #1{#1}          \fi
\ifx \showarticletitle \undefined \def \showarticletitle #1{#1}   \fi
\ifx \showURL      \undefined \def \showURL       {\relax}        \fi
\providecommand\bibfield[2]{#2}
\providecommand\bibinfo[2]{#2}
\providecommand\natexlab[1]{#1}
\providecommand\showeprint[2][]{arXiv:#2}

\bibitem[Adnan et~al\mbox{.}(2022)]%
        {97c9f5c0c4714251a5c377616bf32211}
\bibfield{author}{\bibinfo{person}{Mohammed Adnan}, \bibinfo{person}{Shivam
  Kalra}, \bibinfo{person}{{Jesse C.} Cresswell}, \bibinfo{person}{{Graham W.}
  Taylor}, {and} \bibinfo{person}{{Hamid R.} Tizhoosh}.}
  \bibinfo{year}{2022}\natexlab{}.
\newblock \showarticletitle{Federated learning and differential privacy for
  medical image analysis}.
\newblock \bibinfo{journal}{\emph{Scientific Reports}} \bibinfo{volume}{12},
  \bibinfo{number}{1} (\bibinfo{date}{Dec.} \bibinfo{year}{2022}).
\newblock
\showISSN{2045-2322}
\urldef\tempurl%
\url{https://doi.org/10.1038/s41598-022-05539-7}
\showDOI{\tempurl}


\bibitem[Alansary et~al\mbox{.}(2018)]%
        {alansary2018automatic}
\bibfield{author}{\bibinfo{person}{Amir Alansary}, \bibinfo{person}{Loic
  Le~Folgoc}, \bibinfo{person}{Ghislain Vaillant}, \bibinfo{person}{Ozan
  Oktay}, \bibinfo{person}{Yuanwei Li}, \bibinfo{person}{Wenjia Bai},
  \bibinfo{person}{Jonathan Passerat-Palmbach}, \bibinfo{person}{Ricardo
  Guerrero}, \bibinfo{person}{Konstantinos Kamnitsas},
  \bibinfo{person}{Benjamin Hou}, {et~al\mbox{.}}}
  \bibinfo{year}{2018}\natexlab{}.
\newblock \showarticletitle{Automatic view planning with multi-scale deep
  reinforcement learning agents}. In \bibinfo{booktitle}{\emph{International
  Conference on Medical Image Computing and Computer-Assisted Intervention}}.
  Springer, \bibinfo{pages}{277--285}.
\newblock


\bibitem[Chen et~al\mbox{.}(2019)]%
        {Chen2019AsynchronousOF}
\bibfield{author}{\bibinfo{person}{Yujing Chen}, \bibinfo{person}{Yue Ning},
  {and} \bibinfo{person}{Huzefa Rangwala}.} \bibinfo{year}{2019}\natexlab{}.
\newblock \showarticletitle{Asynchronous Online Federated Learning for Edge
  Devices}.
\newblock \bibinfo{journal}{\emph{ArXiv}}  \bibinfo{volume}{abs/1911.02134}
  (\bibinfo{year}{2019}).
\newblock


\bibitem[Cheng et~al\mbox{.}(2022)]%
        {CHENG2022102313}
\bibfield{author}{\bibinfo{person}{Junlong Cheng}, \bibinfo{person}{Shengwei
  Tian}, \bibinfo{person}{Long Yu}, \bibinfo{person}{Chengrui Gao},
  \bibinfo{person}{Xiaojing Kang}, \bibinfo{person}{Xiang Ma},
  \bibinfo{person}{Weidong Wu}, \bibinfo{person}{Shijia Liu}, {and}
  \bibinfo{person}{Hongchun Lu}.} \bibinfo{year}{2022}\natexlab{}.
\newblock \showarticletitle{ResGANet: Residual group attention network for
  medical image classification and segmentation}.
\newblock \bibinfo{journal}{\emph{Medical Image Analysis}}
  \bibinfo{volume}{76} (\bibinfo{year}{2022}), \bibinfo{pages}{102313}.
\newblock
\showISSN{1361-8415}
\urldef\tempurl%
\url{https://doi.org/10.1016/j.media.2021.102313}
\showDOI{\tempurl}


\bibitem[Huang et~al\mbox{.}(2022)]%
        {2204.13591}
\bibfield{author}{\bibinfo{person}{Yixing Huang}, \bibinfo{person}{Christoph
  Bert}, \bibinfo{person}{Stefan Fischer}, \bibinfo{person}{Manuel Schmidt},
  \bibinfo{person}{Arnd Dörfler}, \bibinfo{person}{Andreas Maier},
  \bibinfo{person}{Rainer Fietkau}, {and} \bibinfo{person}{Florian Putz}.}
  \bibinfo{year}{2022}\natexlab{}.
\newblock \bibinfo{title}{Continual Learning for Peer-to-Peer Federated
  Learning: A Study on Automated Brain Metastasis Identification}.
\newblock
\newblock
\showeprint{arXiv:2204.13591}


\bibitem[Jiang et~al\mbox{.}(2022)]%
        {Jiang_Wang_Dou_2022}
\bibfield{author}{\bibinfo{person}{Meirui Jiang}, \bibinfo{person}{Zirui Wang},
  {and} \bibinfo{person}{Qi Dou}.} \bibinfo{year}{2022}\natexlab{}.
\newblock \showarticletitle{HarmoFL: Harmonizing Local and Global Drifts in
  Federated Learning on Heterogeneous Medical Images}.
\newblock \bibinfo{journal}{\emph{Proceedings of the AAAI Conference on
  Artificial Intelligence}} \bibinfo{volume}{36}, \bibinfo{number}{1}
  (\bibinfo{date}{Jun.} \bibinfo{year}{2022}), \bibinfo{pages}{1087--1095}.
\newblock
\urldef\tempurl%
\url{https://doi.org/10.1609/aaai.v36i1.19993}
\showDOI{\tempurl}


\bibitem[Karani et~al\mbox{.}(2018)]%
        {10.1007/978-3-030-00928-1_54}
\bibfield{author}{\bibinfo{person}{Neerav Karani}, \bibinfo{person}{Krishna
  Chaitanya}, \bibinfo{person}{Christian Baumgartner}, {and}
  \bibinfo{person}{Ender Konukoglu}.} \bibinfo{year}{2018}\natexlab{}.
\newblock \showarticletitle{A Lifelong Learning Approach to Brain MR
  Segmentation Across Scanners and Protocols}. In
  \bibinfo{booktitle}{\emph{Medical Image Computing and Computer Assisted
  Intervention -- MICCAI 2018}},
  \bibfield{editor}{\bibinfo{person}{Alejandro~F. Frangi},
  \bibinfo{person}{Julia~A. Schnabel}, \bibinfo{person}{Christos Davatzikos},
  \bibinfo{person}{Carlos Alberola-L{\'o}pez}, {and} \bibinfo{person}{Gabor
  Fichtinger}} (Eds.). \bibinfo{publisher}{Springer International Publishing},
  \bibinfo{address}{Cham}, \bibinfo{pages}{476--484}.
\newblock
\showISBNx{978-3-030-00928-1}


\bibitem[Khairandish et~al\mbox{.}(2022)]%
        {KHAIRANDISH2022290}
\bibfield{author}{\bibinfo{person}{M.O. Khairandish}, \bibinfo{person}{M.
  Sharma}, \bibinfo{person}{V. Jain}, \bibinfo{person}{J.M. Chatterjee}, {and}
  \bibinfo{person}{N.Z. Jhanjhi}.} \bibinfo{year}{2022}\natexlab{}.
\newblock \showarticletitle{A Hybrid CNN-SVM Threshold Segmentation Approach
  for Tumor Detection and Classification of MRI Brain Images}.
\newblock \bibinfo{journal}{\emph{IRBM}} \bibinfo{volume}{43},
  \bibinfo{number}{4} (\bibinfo{year}{2022}), \bibinfo{pages}{290--299}.
\newblock
\showISSN{1959-0318}
\urldef\tempurl%
\url{https://doi.org/10.1016/j.irbm.2021.06.003}
\showDOI{\tempurl}


\bibitem[Liu et~al\mbox{.}(2022)]%
        {Liu2022AsynchronousDF}
\bibfield{author}{\bibinfo{person}{Qi Liu}, \bibinfo{person}{Bo-Jun Yang},
  \bibinfo{person}{Zhaojian Wang}, \bibinfo{person}{Dafeng Zhu},
  \bibinfo{person}{Xinyi Wang}, \bibinfo{person}{Kai Ma}, {and}
  \bibinfo{person}{Xinping Guan}.} \bibinfo{year}{2022}\natexlab{}.
\newblock \showarticletitle{Asynchronous Decentralized Federated Learning for
  Collaborative Fault Diagnosis of PV Stations}.
\newblock \bibinfo{journal}{\emph{IEEE Transactions on Network Science and
  Engineering}}  \bibinfo{volume}{9} (\bibinfo{year}{2022}),
  \bibinfo{pages}{1680--1696}.
\newblock


\bibitem[Menze et~al\mbox{.}(2014)]%
        {menze2014multimodal}
\bibfield{author}{\bibinfo{person}{Bjoern~H Menze}, \bibinfo{person}{Andras
  Jakab}, \bibinfo{person}{Stefan Bauer}, \bibinfo{person}{Jayashree
  Kalpathy-Cramer}, \bibinfo{person}{Keyvan Farahani}, \bibinfo{person}{Justin
  Kirby}, \bibinfo{person}{Yuliya Burren}, \bibinfo{person}{Nicole Porz},
  \bibinfo{person}{Johannes Slotboom}, \bibinfo{person}{Roland Wiest},
  {et~al\mbox{.}}} \bibinfo{year}{2014}\natexlab{}.
\newblock \showarticletitle{The multimodal brain tumor image segmentation
  benchmark (BRATS)}.
\newblock \bibinfo{journal}{\emph{IEEE transactions on medical imaging}}
  \bibinfo{volume}{34}, \bibinfo{number}{10} (\bibinfo{year}{2014}),
  \bibinfo{pages}{1993--2024}.
\newblock


\bibitem[Mnih et~al\mbox{.}(2013)]%
        {mnih2013playing}
\bibfield{author}{\bibinfo{person}{Volodymyr Mnih}, \bibinfo{person}{Koray
  Kavukcuoglu}, \bibinfo{person}{David Silver}, \bibinfo{person}{Alex Graves},
  \bibinfo{person}{Ioannis Antonoglou}, \bibinfo{person}{Daan Wierstra}, {and}
  \bibinfo{person}{Martin Riedmiller}.} \bibinfo{year}{2013}\natexlab{}.
\newblock \showarticletitle{Playing atari with deep reinforcement learning}.
\newblock \bibinfo{journal}{\emph{arXiv preprint arXiv:1312.5602}}
  (\bibinfo{year}{2013}).
\newblock


\bibitem[Nguyen et~al\mbox{.}(2022)]%
        {PPR:PPR463670}
\bibfield{author}{\bibinfo{person}{TV Nguyen}, \bibinfo{person}{MA Dakka},
  \bibinfo{person}{SM Diakiw}, \bibinfo{person}{MD VerMilyea},
  \bibinfo{person}{M Perugini}, \bibinfo{person}{JMM Hall}, {and}
  \bibinfo{person}{D Perugini}.} \bibinfo{year}{2022}\natexlab{}.
\newblock \bibinfo{title}{A Novel Decentralized Federated Learning Approach to
  Train on Globally Distributed, Poor Quality, and Protected Private Medical
  Data}.
\newblock
\newblock
\urldef\tempurl%
\url{https://doi.org/10.21203/rs.3.rs-1371143/v1}
\showDOI{\tempurl}


\bibitem[Noothout et~al\mbox{.}(2020)]%
        {9139480}
\bibfield{author}{\bibinfo{person}{Julia M.~H. Noothout},
  \bibinfo{person}{Bob~D. De~Vos}, \bibinfo{person}{Jelmer~M. Wolterink},
  \bibinfo{person}{Elbrich~M. Postma}, \bibinfo{person}{Paul A.~M. Smeets},
  \bibinfo{person}{Richard A.~P. Takx}, \bibinfo{person}{Tim Leiner},
  \bibinfo{person}{Max~A. Viergever}, {and} \bibinfo{person}{Ivana Išgum}.}
  \bibinfo{year}{2020}\natexlab{}.
\newblock \showarticletitle{Deep Learning-Based Regression and Classification
  for Automatic Landmark Localization in Medical Images}.
\newblock \bibinfo{journal}{\emph{IEEE Transactions on Medical Imaging}}
  \bibinfo{volume}{39}, \bibinfo{number}{12} (\bibinfo{year}{2020}),
  \bibinfo{pages}{4011--4022}.
\newblock
\urldef\tempurl%
\url{https://doi.org/10.1109/TMI.2020.3009002}
\showDOI{\tempurl}


\bibitem[Pan et~al\mbox{.}(2022)]%
        {PAN2022103824}
\bibfield{author}{\bibinfo{person}{Liangrui Pan}, \bibinfo{person}{Hetian
  Wang}, \bibinfo{person}{Lian Wang}, \bibinfo{person}{Boya Ji},
  \bibinfo{person}{Mingting Liu}, \bibinfo{person}{Mitchai Chongcheawchamnan},
  \bibinfo{person}{Jin Yuan}, {and} \bibinfo{person}{Shaoliang Peng}.}
  \bibinfo{year}{2022}\natexlab{}.
\newblock \showarticletitle{Noise-reducing attention cross fusion learning
  transformer for histological image classification of osteosarcoma}.
\newblock \bibinfo{journal}{\emph{Biomedical Signal Processing and Control}}
  \bibinfo{volume}{77} (\bibinfo{year}{2022}), \bibinfo{pages}{103824}.
\newblock
\showISSN{1746-8094}
\urldef\tempurl%
\url{https://doi.org/10.1016/j.bspc.2022.103824}
\showDOI{\tempurl}


\bibitem[Parekh et~al\mbox{.}(2020)]%
        {parekh2020multitask}
\bibfield{author}{\bibinfo{person}{Vishwa~S Parekh}, \bibinfo{person}{Vladimir
  Braverman}, \bibinfo{person}{Michael~A Jacobs}, {et~al\mbox{.}}}
  \bibinfo{year}{2020}\natexlab{}.
\newblock \showarticletitle{Multitask radiological modality invariant landmark
  localization using deep reinforcement learning}. In
  \bibinfo{booktitle}{\emph{Medical Imaging with Deep Learning}}. PMLR,
  \bibinfo{pages}{588--600}.
\newblock


\bibitem[Rieke et~al\mbox{.}(2020)]%
        {58f2965c5c4847d8b5e02e9e4408799d}
\bibfield{author}{\bibinfo{person}{Nicola Rieke}, \bibinfo{person}{Jonny
  Hancox}, \bibinfo{person}{Wenqi Li}, \bibinfo{person}{Fausto Milletar{\`i}},
  \bibinfo{person}{{Holger R.} Roth}, \bibinfo{person}{Shadi Albarqouni},
  \bibinfo{person}{Spyridon Bakas}, \bibinfo{person}{{Mathieu N.} Galtier},
  \bibinfo{person}{{Bennett A.} Landman}, \bibinfo{person}{Klaus Maier-Hein},
  \bibinfo{person}{S{\'e}bastien Ourselin}, \bibinfo{person}{Micah Sheller},
  \bibinfo{person}{{Ronald M.} Summers}, \bibinfo{person}{Andrew Trask},
  \bibinfo{person}{Daguang Xu}, \bibinfo{person}{Maximilian Baust}, {and}
  \bibinfo{person}{{M. Jorge} Cardoso}.} \bibinfo{year}{2020}\natexlab{}.
\newblock \showarticletitle{The future of digital health with federated
  learning}.
\newblock \bibinfo{journal}{\emph{npj Digital Medicine}} \bibinfo{volume}{3},
  \bibinfo{number}{1} (\bibinfo{date}{1 Dec.} \bibinfo{year}{2020}).
\newblock
\showISSN{2398-6352}
\urldef\tempurl%
\url{https://doi.org/10.1038/s41746-020-00323-1}
\showDOI{\tempurl}


\bibitem[Rolnick et~al\mbox{.}(2019)]%
        {rolnick2019experience}
\bibfield{author}{\bibinfo{person}{David Rolnick}, \bibinfo{person}{Arun
  Ahuja}, \bibinfo{person}{Jonathan Schwarz}, \bibinfo{person}{Timothy
  Lillicrap}, {and} \bibinfo{person}{Gregory Wayne}.}
  \bibinfo{year}{2019}\natexlab{}.
\newblock \showarticletitle{Experience replay for continual learning}.
\newblock \bibinfo{journal}{\emph{Advances in Neural Information Processing
  Systems}}  \bibinfo{volume}{32} (\bibinfo{year}{2019}).
\newblock


\bibitem[Roth et~al\mbox{.}(2020)]%
        {10.1007/978-3-030-60548-3_18}
\bibfield{author}{\bibinfo{person}{Holger~R. Roth}, \bibinfo{person}{Ken
  Chang}, \bibinfo{person}{Praveer Singh}, \bibinfo{person}{Nir Neumark},
  \bibinfo{person}{Wenqi Li}, \bibinfo{person}{Vikash Gupta},
  \bibinfo{person}{Sharut Gupta}, \bibinfo{person}{Liangqiong Qu},
  \bibinfo{person}{Alvin Ihsani}, \bibinfo{person}{Bernardo~C. Bizzo},
  \bibinfo{person}{Yuhong Wen}, \bibinfo{person}{Varun Buch},
  \bibinfo{person}{Meesam Shah}, \bibinfo{person}{Felipe Kitamura},
  \bibinfo{person}{Matheus Mendon\c{c}a}, \bibinfo{person}{Vitor Lavor},
  \bibinfo{person}{Ahmed Harouni}, \bibinfo{person}{Colin Compas},
  \bibinfo{person}{Jesse Tetreault}, \bibinfo{person}{Prerna Dogra},
  \bibinfo{person}{Yan Cheng}, \bibinfo{person}{Selnur Erdal},
  \bibinfo{person}{Richard White}, \bibinfo{person}{Behrooz Hashemian},
  \bibinfo{person}{Thomas Schultz}, \bibinfo{person}{Miao Zhang},
  \bibinfo{person}{Adam McCarthy}, \bibinfo{person}{B.~Min Yun},
  \bibinfo{person}{Elshaimaa Sharaf}, \bibinfo{person}{Katharina~V. Hoebel},
  \bibinfo{person}{Jay~B. Patel}, \bibinfo{person}{Bryan Chen},
  \bibinfo{person}{Sean Ko}, \bibinfo{person}{Evan Leibovitz},
  \bibinfo{person}{Etta~D. Pisano}, \bibinfo{person}{Laura Coombs},
  \bibinfo{person}{Daguang Xu}, \bibinfo{person}{Keith~J. Dreyer},
  \bibinfo{person}{Ittai Dayan}, \bibinfo{person}{Ram~C. Naidu},
  \bibinfo{person}{Mona Flores}, \bibinfo{person}{Daniel Rubin}, {and}
  \bibinfo{person}{Jayashree Kalpathy-Cramer}.}
  \bibinfo{year}{2020}\natexlab{}.
\newblock \showarticletitle{Federated Learning for Breast Density
  Classification: A Real-World Implementation}. In
  \bibinfo{booktitle}{\emph{Domain Adaptation and Representation Transfer, and
  Distributed and Collaborative Learning: Second MICCAI Workshop, DART 2020,
  and First MICCAI Workshop, DCL 2020, Held in Conjunction with MICCAI 2020,
  Lima, Peru, October 4–8, 2020, Proceedings}}.
  \bibinfo{publisher}{Springer-Verlag}, \bibinfo{address}{Berlin, Heidelberg},
  \bibinfo{pages}{181–191}.
\newblock


\bibitem[Roy et~al\mbox{.}(2019)]%
        {Roy2019BrainTorrentAP}
\bibfield{author}{\bibinfo{person}{Abhijit~Guha Roy}, \bibinfo{person}{Shayan
  Siddiqui}, \bibinfo{person}{Sebastian P{\"o}lsterl}, \bibinfo{person}{Nassir
  Navab}, {and} \bibinfo{person}{Christian Wachinger}.}
  \bibinfo{year}{2019}\natexlab{}.
\newblock \showarticletitle{BrainTorrent: A Peer-to-Peer Environment for
  Decentralized Federated Learning}.
\newblock \bibinfo{journal}{\emph{ArXiv}}  \bibinfo{volume}{abs/1905.06731}
  (\bibinfo{year}{2019}).
\newblock


\bibitem[Tripathi et~al\mbox{.}(2022)]%
        {doi:10.1080/02564602.2021.1937349}
\bibfield{author}{\bibinfo{person}{Sumit Tripathi},
  \bibinfo{person}{Taresh~Sarvesh Sharan}, \bibinfo{person}{Shiru Sharma},
  {and} \bibinfo{person}{Neeraj Sharma}.} \bibinfo{year}{2022}\natexlab{}.
\newblock \showarticletitle{An Augmented Deep Learning Network with Noise
  Suppression Feature for Efficient Segmentation of Magnetic Resonance Images}.
\newblock \bibinfo{journal}{\emph{IETE Technical Review}} \bibinfo{volume}{39},
  \bibinfo{number}{4} (\bibinfo{year}{2022}), \bibinfo{pages}{960--973}.
\newblock
\urldef\tempurl%
\url{https://doi.org/10.1080/02564602.2021.1937349}
\showDOI{\tempurl}


\bibitem[Vlontzos et~al\mbox{.}(2019)]%
        {vlontzos2019multiple}
\bibfield{author}{\bibinfo{person}{Athanasios Vlontzos}, \bibinfo{person}{Amir
  Alansary}, \bibinfo{person}{Konstantinos Kamnitsas}, \bibinfo{person}{Daniel
  Rueckert}, {and} \bibinfo{person}{Bernhard Kainz}.}
  \bibinfo{year}{2019}\natexlab{}.
\newblock \showarticletitle{Multiple Landmark Detection using Multi-Agent
  Reinforcement Learning}. In \bibinfo{booktitle}{\emph{International
  Conference on Medical Image Computing and Computer-Assisted Intervention}}.
  Springer, \bibinfo{pages}{262--270}.
\newblock


\bibitem[Wang et~al\mbox{.}(2022)]%
        {electronics11101548}
\bibfield{author}{\bibinfo{person}{Zhao Wang}, \bibinfo{person}{Yifan Hu},
  \bibinfo{person}{Shiyang Yan}, \bibinfo{person}{Zhihao Wang},
  \bibinfo{person}{Ruijie Hou}, {and} \bibinfo{person}{Chao Wu}.}
  \bibinfo{year}{2022}\natexlab{}.
\newblock \showarticletitle{Efficient Ring-Topology Decentralized Federated
  Learning with Deep Generative Models for Medical Data in eHealthcare
  Systems}.
\newblock \bibinfo{journal}{\emph{Electronics}} \bibinfo{volume}{11},
  \bibinfo{number}{10} (\bibinfo{year}{2022}).
\newblock
\showISSN{2079-9292}
\urldef\tempurl%
\url{https://doi.org/10.3390/electronics11101548}
\showDOI{\tempurl}


\bibitem[Yan et~al\mbox{.}(2021)]%
        {9268161}
\bibfield{author}{\bibinfo{person}{Zengqiang Yan}, \bibinfo{person}{Jeffry
  Wicaksana}, \bibinfo{person}{Zhiwei Wang}, \bibinfo{person}{Xin Yang}, {and}
  \bibinfo{person}{Kwang-Ting Cheng}.} \bibinfo{year}{2021}\natexlab{}.
\newblock \showarticletitle{Variation-Aware Federated Learning With
  Multi-Source Decentralized Medical Image Data}.
\newblock \bibinfo{journal}{\emph{IEEE Journal of Biomedical and Health
  Informatics}} \bibinfo{volume}{25}, \bibinfo{number}{7}
  (\bibinfo{year}{2021}), \bibinfo{pages}{2615--2628}.
\newblock
\urldef\tempurl%
\url{https://doi.org/10.1109/JBHI.2020.3040015}
\showDOI{\tempurl}


\bibitem[Yoo et~al\mbox{.}(2021)]%
        {10.1007/978-3-030-91387-8_1}
\bibfield{author}{\bibinfo{person}{Joo~Hun Yoo}, \bibinfo{person}{Hyejun
  Jeong}, \bibinfo{person}{Jaehyeok Lee}, {and} \bibinfo{person}{Tai-Myoung
  Chung}.} \bibinfo{year}{2021}\natexlab{}.
\newblock \showarticletitle{Federated Learning: Issues in Medical Application}.
  In \bibinfo{booktitle}{\emph{Future Data and Security Engineering: 8th
  International Conference, FDSE 2021, Virtual Event, November 24–26, 2021,
  Proceedings}}. \bibinfo{publisher}{Springer}, \bibinfo{address}{Berlin,
  Heidelberg}, \bibinfo{pages}{3–22}.
\newblock


\end{thebibliography}

\appendix

\section{Research Method}

\subsection{Deep Reinforcement Learning}
We created a deep reinforcement learning framework that utilizes the deep Q-network (DQN) algorithm, which is depicted in Figure \ref{fig2}. The 3D DQN model we used in this paper was adapted from existing works \cite{mnih2013playing,alansary2018automatic,vlontzos2019multiple,parekh2020multitask}. The environment was represented by a 3D imaging volume with x, y, and z dimensions. The agent was represented by a 3-dimensional bounding box with six possible actions: moving in the positive or negative in the x, y, or z axis. The state was defined by the current location (or a chain of locations) of the agent, each represented by a 3-dimensional bounding box. The reward was calculated by the change in distance to the target landmark location before and after the agent takes an action. The agent's exploration within the environment generated state-action-reward-resulting state $[s,a,r,s']$ tuples, which were recorded in the experience replay buffer (ERB) over multiple  episodes. The information contained in the ERBs are non-sensitive information, as the action and reward are numbers regarding the DRL model, and the state and resulting states are small fractions of the total 3D image, roughly $0.3\%$.

\subsection{Lifelong Learning}
We implemented lifelong learning using selective experience replay \cite{rolnick2019experience}. The goal of selective experience replay is to avoid catastrophic forgetting by focusing on selected experiences from previous tasks. Additionally, this technique is agnostic to the model being used and enables the sharing of experiences across different models. To achieve lifelong learning, we utilized a selective experience replay buffer that collects a sequence of experience samples throughout the model's training process. In order to learn a generalized representation of both current and past tasks, the model selects a batch of experiences from both the ERB of its current task and from the replay buffers of previous tasks during training.

\subsection{Asynchronous Decentralized Federated Lifelong Learning}
We developed the Asynchronous Decentralized Federated lifelong Learning (ADFLL) by constructing a network of lifelong deep reinforcement learning agents. Each agent shares their database of personal experiences with each other to facilitate learning from each other experiences. More specifically, once an agent finishes training with a dataset and an ERB, the resulting experience from the training is shared with the network. Furthermore, we modified the training setup for each agent to sample experiences from the current dataset ERB, the agent's personal experiences and the incoming experiences from other agents, as shown in Fig.~\ref{fig2}. As a result, every agent in the network can learn from each other's experiences, thereby integrating federated lifelong learning capability. 

\begin{figure}[htb]
\centering{}
\includegraphics[width=0.5\textwidth]{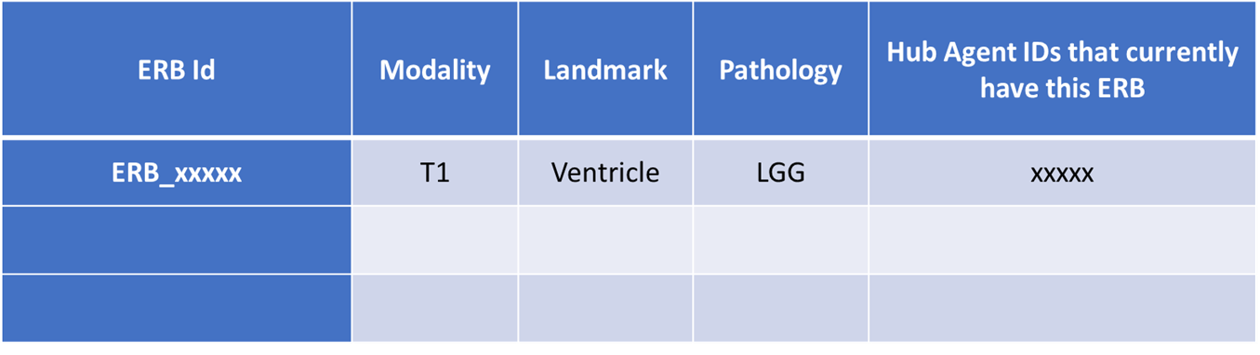}
\caption{Snapshot of the shared database maintained by the hub nodes} 
\label{fig7}
\end{figure}

In a naive setup, every agent would communicate their experiences with every other agent in the network. However, such an all-to-all communication setup is highly inefficient and not scalable as it would require a large amount of communication bandwidth. To address this issue, we implemented a homogeneous distributed database system as illustrated in Fig.~\ref{fig2}. As shown in Fig.~\ref{fig2}, our network consists of a predefined set of hub nodes that act as communication hubs for spatially adjacent nodes in the network. Subsequently, every agent in the network exclusively communicates with their nearest hub node at the end of each personal training round. The experience sharing between an agent and a hub node is bidirectional. Finally, every hub node maintains a shared experience database on the network as shown in Fig.~\ref{fig7}. The hub nodes periodically communicate with each other to synchronize their databases. The agents in the system are not required to have standardized training speed or start training at the same time. The hub will regulate and preserve the experiences in the system and agents in the system can train on different tasks. An example of this system is demonstrated in Fig.~\ref{fig2}.

The advantage of our system setup is that it is robust against node or hub failures. When a node fails, the only loss is the training information from that node, and when a hub fails, the loss is the ERBs it contains but other hubs do not. Moreover, the communication complexity is linear with respect to the number of nodes, each node only needs to communicate with its respective hub, and hubs sync periodically. Compared to other federated learning systems, centralized or not, they either are prone to system-wide failure caused by a node failure, or sacrifice communication complexity to prevent system-wide failures.

\end{document}